# BioBERT Based SNP-traits Associations Extraction from Biomedical Literature

Mohammad Dehghani*
School of Electrical and Computer Engineering,
University of Tehran,
Tehran, Iran
dehghani.mohammad@ut.ac.ir

Behrouz Bokharaeian
Amol University of Special Modern Technologies,
Mazandaran, Iran
bokharaeian@gmail.com

Zahra Yazdanparast
School of Electrical and Computer Engineering,
Tarbiat Modares University,
Tehran, Iran
zahra.yazdanparast@modares.ac.ir

*Abstract*—Scientific literature contains a considerable amount of information that provides an excellent opportunity for developing text mining methods to extract biomedical relationships. An important type of information is the relationship between singular nucleotide polymorphisms (SNP) and traits. In this paper, we present a BioBERT-GRU method to identify SNP- traits associations. Based on the evaluation of our method on the SNPPhenA dataset, it is concluded that this new method performs better than previous machine learning and deep learning based methods. BioBERT-GRU achieved the result a precision of 0.883, recall of 0.882 and F1-score of 0.881.

## I. INTRODUCTION

In a genome-wide association study (GWAS), genetic variants across genomes are analyzed to identify those that are statistically related to a particular disease or trait. GWAS are designed to identify associations between genotypes and phenotypes [1]. They examine differences in the allele frequency of genetic variants in individuals that are genetically related yet differ phenotypically. The most common genetic variants studied in GWAS are single- nucleotide polymorphisms (SNPs) [2]. SNP is a single-base mutation at the DNA level [3]. These polymorphisms, which are located near virtually every gene, can be used as genetic markers. It is also possible to detect associations between genes and phenotypes using SNPs, particularly in diseases that have multifactorial genetics [4].

Scientific text based publications can provide valuable insights to identify associations that will be important for future scientific discoveries. Although reading, interpreting, and analyzing rapidly increasing volumes of texts manually is very time-consuming and beyond anyone's capabilities [5]. Therefore, it is essential to develop tools that collect relevant information from the literature and integrate it with existing biomedical knowledge [6]. This problem can be addressed by using natural language processing (NLP) and text mining techniques to automatically process biomedical literature in order to extract named entities and associations [7].

Biomedical text mining has become an important component of scientific research because of applications such as relation extraction [8], identification of bio-events [9], hypothesis generation [10]. Biomedical researchers have used a variety of approaches for relation extraction, including co-occurrence-based [11], rule-based [12], machine learning [13], and natural language processing [14]. Development of neural network models, including deep learning, convolutional neural networks (CNNs), and recurrent neural networks (RNNs) has led to new methods of identifying relationships [15]. Recently, hybrid approaches have been proposed to achieve better performance by combining two or more methods [16].

In this paper, we present some methods based on deep learning approach called BioBERT-GRU to extract SNPs and traits associations from biomedical literature using SNPPhenA corpus.

The rest of the paper is organized as follows: In Section 2 related works are reviewed, Section 3 presents our proposed model, Sections 4 discusses the results, and finally, Section 5 concludes the paper.

## II. RELATED WORKS

Biomedicine studies have used text mining in different applications such as identification of drug-drug interactions, drug-indications, drug-gene interactions, and gene-disease.
[17] used biomedical literature from Wikipedia and DrugBank to extract drug-drug interactions (DDI) using BioBERT and BiGRU. The main contribution of their study was the consideration of drug names in the DDIs corpus. For the extraction of DDIs, [18] applied SGRU-CNN model to lexical data and entity position data while trying to achieve a balance between a simpler method and a higher level of model performance. The study by [19] examined a list of features for identifying neutral candidates, as well as features extracted from negation cues for detecting drug-drug interactions. The authors of [20] present a text mining method for retrieving DDI and ADR (Adverse Drug Reaction) information related to the patient population, gender, pharmacokinetics, and pharmacodynamics from PubMed. [21] developed a SVM model combining three distinct corpora (ADE, Twitter, and DailyStrength2) and found that combining corpora from similar sources improved the detection of ADR. A complex method of detecting ADR from twitter by using hierarchical

tweet representations and multi-head self-attention has been developed by [22]. In order to improve ADR detection performance when only a limited amount of annotation is available for training, [23] proposed a framework based on adversarial transfer learning.

PubMed and PubMed Central were used by [24] to identify search terms related to dry mouth and/or xerostomia. Using text-mining tools, they investigated xerostomia and dry mouth concepts, identified molecular-genes interactions, and assessed the effects of drugs on genes and pathways. The primary objective of the study [25] was to identify genes related to breast cancer and periodontitis by using text mining, as well as identify drug candidates targeting these genes.

In [26], the cosine similarity between gene vectors and disease vectors was applied to identify gene-disease associations. In their approach, MeSH (Medical Subject Headings) is integrated with term weights and co-occurrence techniques. Using an ensemble support vector machine (ensemble SVM), [27] analyze four gold standard corpora to extract gene-disease associations. Their feature set consisting of conceptual, syntax and semantic properties that have been jointly learned through word embedding (via Word2Vec). Several other methods have been proposed for extracting gene-disease associations from single sentences, abstract texts, or full-text articles, such as Be-Free [6], DTMiner [28], BioBERT [29], and RENET [30]. Recent research by [31] proposes RENET2 and extracts over three million associations between genes and diseases from PMC. The RENET2 algorithm is based on deep learning and employs Section Filtering and ambiguous relation modeling to extract associations from full-text articles.

The authors of [32] presented a pipeline for identifying associations between genes and phenotypes of Autism Spectrum Disorder, based on NLP methods using articles from PubMed Central (PMC). A self-training algorithm was developed by [33] to increase the training set, as well as a machine-learned model to detect genotype-phenotype associations from biomedical text.

## III. PROPOSED METHOD

An overview of the proposed model is shown in Fig. 1. We used the SNPhenA corpus as input data. The preprocessing process involves tokenization, removal of stop words, stemming, and lemmatization using the NLTK and Spacy libraries. The SNP-Phenotype association candidates were classified into positive and negative/neutral using proposed classifier.

### A. CNN

A CNN is one of the most widely used algorithms in deep learning and has been extensively applied to computer vision [34], NLP [35], and speech processing [36]. CNN models use convolutional layers to connect a subset of inputs to their preceding layers. Pooling layers in CNN reduce the parameters in a network by downsampling each feature map, thus speeding up the training process and reducing overfitting. A number of Pooling techniques are available, however, max-Pooling is most commonly used in which the pooling window contains maximum values. The final layer of a CNN is typically fully connected (e.g., Softmax), that determines the probability that an instance belong to a particular class [37].

### B. GRU

The model we propose uses GRU, which is a specific kind of recurrent neural network (RNN). GRU is very similar to LSTM, with less complicated structure. It has an update z and a reset gate r which decide what information should be passed to the output. Equations (1-4) explain the GRU gates and cells [38]. We used Biderectional GRU to train our model.

Update gate $\quad z = \sigma (W_z h_{t-1} + U_z x_t) \quad$ (1)

Reset gate $\quad r = \sigma (W_r h_{t-1} + U_r x_t) \quad$ (2)

Cell state $\quad C = tanh (W_c (h_{t-1} * r) + U_c x_t) \quad$ (3)

New state $\quad h_t = (z * c) + ((1 - z) * h_{t-1}) \quad$ (4)

### C. BioBERT

BERT stands for bidirectional encoder representations from transformers [39]. BERT focuses on jointly conditioning the left and right context in all layers in order to pre-train deep bidirectional representations. Consequently, these representations can be fine-tuned to produce state-of-the-art models for a wide range of NLP tasks without substantial modifications [40].

There are many domain-specific nouns and terms in biomedical texts that are primarily understood by biomedical researchers. BioBERT is a representation model trained on large-scale biomedical corpora such as PubMed abstracts and PubMed Central full-text articles [29].

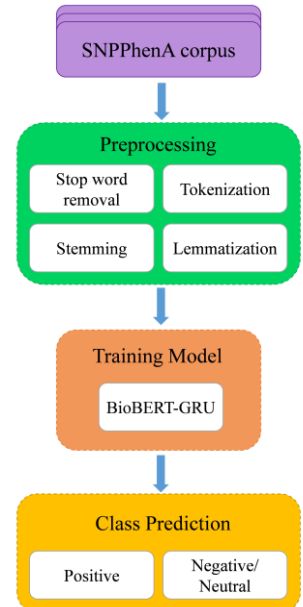

Fig. 1. An overview of proposed model.

Fig. 2 shows proposed architecture. pre-trained BioBERT model is used for feature representation followed by CNN layer for feature extraction. Next, we used maxpooling. Our model is trained using bidirectional GRU. Then, a fully connected layer with RELU activation function was added, followed by a drop out layer. Finaly we have fully connected layer with Softmax activation function that generate the output. The output can be one of these two classes: positive or negative/nutral.

IV. RESULTS

*A. Dataset*

The SNPPhenA dataset [41] contains ranked associations between mutations and traits based on GWA studies. As part of the process, relevant abstracts are collected, SNP-trait associations are identified, negation, modality markers and their level of confidence is determined, etc. In the corpus, associations were classified as positive, negative, or neutral candidates. Positive candidates have clearly indicated associations. Negative candidates have no association. As for the neutral candidates, no clear evidence was found to support an association between mutations and traits. SNPPhenA corpus contain 360 files with 2525 Sentences. There is 811 positive candidates, 325 negative candidates and 180 neutral candidates.

*B. Evaluation metrics*

**Recall** is a measure of how well positive observations are classified. This is defined as the fraction of True Positive out of the total number of positive observations.

$$Recall = \frac{True\ Positive}{True\ Positive + False\ Negative} \quad (5)$$

**Precision** of a classification is determined by the ratio of correctly classified positive observations to all positive observations.

$$Precission = \frac{True\ Positive}{True\ Positive + False\ Positive} \quad (6)$$

**F-score** combines precision and recall and is commonly used in the case of imbalanced data.

$$F1\_score = \frac{2 * Precission * Recall}{Precission + Recall} \quad (7)$$

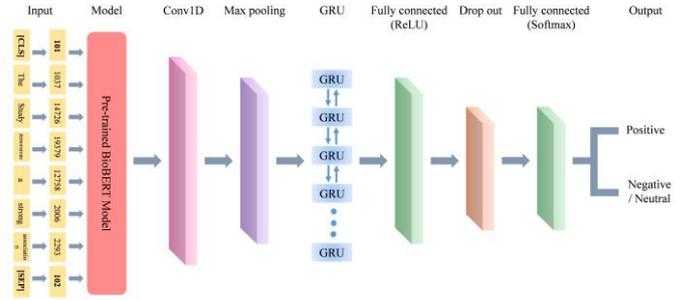

Fig. 2. Proposed BioBERT-GRU architecture.

*C. Evaluation*

To extract SNP-phenotype associations, we used SNPPhenA corups dataset. The data is divided into train data for training models and test data for assessing model performance. It is gathered in different item types including sentence and abstract. We classified the dataset into two classes (positive and negative/neutral) and evaluate our method. Table I shows training parameter of model with the batch size of 16 and 30 epochs. Maximum length of sentences is 70 and for abstract 300 characters. We use Adam optimizer with leraning rate of 1e-4 and epsilon of 1e-7.

Table II shows the results of identifying associations between SNPs and phenotypes at the sentence level. Positive candidates were considered as one class, and negative/neutral candidates as the second class. Precision, recall and F1-score of new model are equal to 88%, which shows better performance in compare to previouse models.

The results of SNP-Phenotype associations detection at the abstract level are shown in Table III. BioBERT-GRU has precision of 0.68, recall of 0.65, and F1-score of 0.64 which shows a significant improvement. It is important to note that although precision is 4% less than PubMedBERT-LSTM model, recall has increased by 6% and F1-score has increased by 12%.

According to the results, it was confirmed that the our new method perform better then existing methods at extracting associations in both sentence and abstract level.

TABLE I. Training parameters.

| Parameter | | Value |
|---|---|---|
| Batch_size | | 16 |
| Epochs | | 30 |
| Max_lenght | Sentence | 70 |
| | Abstract | 300 |
| Adam | leraning rate | 1e-4 |
| | epsilon | 1e-7 |

TABLE II. Results of identifying SNP-Phenotype associations for the test corpus at the sentence level.

| Model Name | Precision | Recall | F1 |
|---|---|---|---|
| CNN-LSTM [42] | 0.732 | 0.723 | 0.723 |
| BERT-LSTM [42] | 0.739 | 0.803 | 0.730 |
| PubMedBERT-LSTM [42] | 0.867 | 0.870 | 0.866 |
| Proposed BioBERT-GRU | **0.883** | **0.882** | **0.881** |

TABLE III. Results of identifying SNP-Phenotype associations for the test corpus at the abstract level.

| Model Name | Precision | Recall | F1 |
|---|---|---|---|
| CNN-LSTM [42] | 0.537 | 0.477 | 0.345 |
| BERT-LSTM [42] | 0.548 | 0.509 | 0.465 |
| PubMedBERT-LSTM [42] | 0.720 | 0.580 | 0.523 |
| Proposed BioBERT-GRU | **0.680** | **0.652** | **0.645** |

## V. CONCLUSION

This paper presents a deep learning based method for extracting SNP-phenotype associations from biomedical literature. We evaluated the models using the SNPPhenA corpus, which can be divided into two classes. Compared to other models, BioBERT-GRU performs better in both sentence and abstract level with F1-score equal to 0.88 and 0.64.

We can determine the type of relationship for future works by determining whether it is a direct relationship or indicative of a pathway effect. Additionally, fuzzy relations can be used as an alternative to crisp relations.

## REFERENCES

[1] V. Tam, N. Patel, M. Turcotte, Y. Bossé, G. Paré, and D. Meyre, "Benefits and limitations of genome-wide association studies," Nature Reviews Genetics, vol. 20, no. 8, pp. 467-484, 2019.
[2] E. Uffelmann et al., "Genome-wide association studies," Nature Reviews Methods Primers, vol. 1, no. 1, p. 59, 2021.
[3] G. T. Marth et al., "A general approach to single-nucleotide polymorphism discovery," Nature genetics, vol. 23, no. 4, pp. 452-456, 1999.
[4] A. Rafalski, "Applications of single nucleotide polymorphisms in crop genetics," Current opinion in plant biology, vol. 5, no. 2, pp. 94-100, 2002.
[5] V. D. Badal et al., "Challenges in the construction of knowledge bases for human microbiome-disease associations," Microbiome, vol. 7, no. 1, pp. 1-15, 2019.
[6] À. Bravo, J. Piñero, N. Queralt-Rosinach, M. Rautschka, and L. I. Furlong, "Extraction of relations between genes and diseases from text and large-scale data analysis: implications for translational research," BMC bioinformatics, vol. 16, pp. 1-17, 2015.
[7] H. Kilicoglu, "Biomedical text mining for research rigor and integrity: tasks, challenges, directions," Briefings in bioinformatics, vol. 19, no. 6, pp. 1400-1414, 2018.
[8] G. Murugesan, S. Abdulkadhar, and J. Natarajan, "Distributed smoothed tree kernel for protein-protein interaction extraction from the biomedical literature," PLoS One, vol. 12, no. 11, p. e0187379, 2017.
[9] R. Harpaz et al., "Text mining for adverse drug events: the promise, challenges, and state of the art," Drug safety, vol. 37, pp. 777-790, 2014.
[10] S. Karimi, C. Wang, A. Metke-Jimenez, R. Gaire, and C. Paris, "Text and data mining techniques in adverse drug reaction detection," ACM Computing Surveys (CSUR), vol. 47, no. 4, pp. 1-39, 2015.
[11] E. S. Chen, G. Hripcsak, H. Xu, M. Markatou, and C. Friedman, "Automated acquisition of disease–drug knowledge from biomedical and clinical documents: an initial study," Journal of the American Medical Informatics Association, vol. 15, no. 1, pp. 87-98, 2008.
[12] K. Ravikumar, M. Rastegar-Mojarad, and H. Liu, "BELMiner: adapting a rule-based relation extraction system to extract biological expression language statements from biomedical literature evidence sentences," Database, vol. 2017, 2017.
[13] A. W. Muzaffar, F. Azam, and U. Qamar, "A relation extraction framework for biomedical text using hybrid feature set," Computational and mathematical methods in medicine, vol. 2015, 2015.
[14] T. C. Rindflesch and M. Fiszman, "The interaction of domain knowledge and linguistic structure in natural language processing: interpreting hypernymic propositions in biomedical text," Journal of biomedical informatics, vol. 36, no. 6, pp. 462-477, 2003.
[15] A. Akkasi and M.-F. Moens, "Causal relationship extraction from biomedical text using deep neural models: A comprehensive survey," Journal of Biomedical Informatics, vol. 119, p. 103820, 2021.
[16] Y. Peng, A. Rios, R. Kavuluru, and Z. Lu, "Chemical-protein relation extraction with ensembles of SVM, CNN, and RNN models," arXiv preprint arXiv:1802.01255, 2018.
[17] Y. Zhu, L. Li, H. Lu, A. Zhou, and X. Qin, "Extracting drug-drug interactions from texts with BioBERT and multiple entity-aware attentions," Journal of biomedical informatics, vol. 106, p. 103451, 2020.

[18] H. Wu et al., "Drug-drug interaction extraction via hybrid neural networks on biomedical literature," Journal of biomedical informatics, vol. 106, p. 103432, 2020.
[19] B. Bokharaeian, A. Diaz, and H. Chitsaz, "Enhancing extraction of drug-drug interaction from literature using neutral candidates, negation, and clause dependency," PLoS One, vol. 11, no. 10, p. e0163480, 2016.
[20] M. S. A. Shukkoor, M. T. H. Baharuldin, and K. Raja, "A Text Mining Protocol for Extracting Drug–Drug Interaction and Adverse Drug Reactions Specific to Patient Population, Pharmacokinetics, Pharmacodynamics, and Disease," in Biomedical Text Mining: Springer, 2022, pp. 259-282.
[21] A. Sarker and G. Gonzalez, "Portable automatic text classification for adverse drug reaction detection via multi-corpus training," Journal of biomedical informatics, vol. 53, pp. 196-207, 2015.
[22] C. Wu, F. Wu, J. Liu, S. Wu, Y. Huang, and X. Xie, "Detecting tweets mentioning drug name and adverse drug reaction with hierarchical tweet representation and multi-head self-attention," in Proceedings of the 2018 EMNLP workshop SMM4H: the 3rd social media mining for health applications workshop & shared task, 2018, pp. 34-37.
[23] Z. Li, Z. Yang, L. Luo, Y. Xiang, and H. Lin, "Exploiting adversarial transfer learning for adverse drug reaction detection from texts," Journal of biomedical informatics, vol. 106, p. 103431, 2020.
[24] M. F. Beckman, E. J. Brennan, C. K. Igba, M. T. Brennan, F. B. Mougeot, and J.-L. C. Mougeot, "A Computational Text Mining-Guided Meta-Analysis Approach to Identify Potential Xerostomia Drug Targets," Journal of Clinical Medicine, vol. 11, no. 5, p. 1442, 2022.
[25] L. Luo, W. Zheng, C. Chen, and S. Sun, "Searching for essential genes and drug discovery in breast cancer and periodontitis via text mining and bioinformatics analysis," Anti-Cancer Drugs, vol. 32, no. 10, p. 1038, 2021.
[26] J. Zhou and B.-q. Fu, "The research on gene-disease association based on text-mining of PubMed," Bmc Bioinformatics, vol. 19, pp. 1-8, 2018.
[27] B. Bhasuran and J. Natarajan, "Automatic extraction of gene-disease associations from literature using joint ensemble learning," PloS one, vol. 13, no. 7, p. e0200699, 2018.
[28] D. Xu et al., "DTMiner: identification of potential disease targets through biomedical literature mining," Bioinformatics, vol. 32, no. 23, pp. 3619-3626, 2016.
[29] J. Lee et al., "BioBERT: a pre-trained biomedical language representation model for biomedical text mining," Bioinformatics, vol. 36, no. 4, pp. 1234-1240, 2020.
[30] Y. Wu, R. Luo, H. C. Leung, H.-F. Ting, and T.-W. Lam, "Renet: A deep learning approach for extracting gene-disease associations from literature," in Research in Computational Molecular Biology: 23rd Annual International Conference, RECOMB 2019, Washington, DC, USA, May 5-8, 2019, Proceedings 23, 2019: Springer, pp. 272-284.
[31] J. Su, Y. Wu, H.-F. Ting, T.-W. Lam, and R. Luo, "RENET2: high-performance full-text gene–disease relation extraction with iterative training data expansion," NAR Genomics and Bioinformatics, vol. 3, no. 3, p. lqab062, 2021.
[32] S. Li, Z. Guo, J. B. Ioffe, Y. Hu, Y. Zhen, and X. Zhou, "Text mining of gene–phenotype associations reveals new phenotypic profiles of autism-associated genes," Scientific Reports, vol. 11, no. 1, pp. 1-12, 2021.
[33] M. Khordad and R. E. Mercer, "Identifying genotype-phenotype relationships in biomedical text," Journal of biomedical semantics, vol. 8, pp. 1-16, 2017.
[34] A. Krizhevsky, I. Sutskever, and G. E. Hinton, "Imagenet classification with deep convolutional neural networks," Communications of the ACM, vol. 60, no. 6, pp. 84-90, 2017.
[35] M. Gimenez, J. Palanca, and V. Botti, "Semantic-based padding in convolutional neural networks for improving the performance in natural language processing. A case of study in sentiment analysis," Neurocomputing, vol. 378, pp. 315-323, 2020.
[36] J.-T. Huang, J. Li, and Y. Gong, "An analysis of convolutional neural networks for speech recognition," in 2015 IEEE International Conference on Acoustics, Speech and Signal Processing (ICASSP), 2015: IEEE, pp. 4989-4993.
[37] S. Pouyanfar et al., "A survey on deep learning: Algorithms, techniques, and applications," ACM Computing Surveys (CSUR), vol. 51, no. 5, pp. 1-36, 2018.
[38] P. T. Yamak, L. Yujian, and P. K. Gadosey, "A comparison between arima, lstm, and gru for time series forecasting," in Proceedings of the 2019 2nd international conference on algorithms, computing and artificial intelligence, 2019, pp. 49-55.
[39] J. Devlin, M.-W. Chang, K. Lee, and K. Toutanova, "Bert: Pre-training of deep bidirectional transformers for language understanding," arXiv preprint arXiv:1810.04805, 2018.
[40] L. Khan, A. Amjad, K. M. Afaq, and H.-T. Chang, "Deep sentiment analysis using CNN-LSTM architecture of English and Roman Urdu text shared in social media," Applied Sciences, vol. 12, no. 5, p. 2694, 2022.
[41] B. Bokharaeian, A. Diaz, N. Taghizadeh, H. Chitsaz, and R. Chavoshinejad, "SNPPhenA: a corpus for extracting ranked associations of single-nucleotide polymorphisms and phenotypes from literature," Journal of biomedical semantics, vol. 8, pp. 1-13, 2017.
[42] B. Bokharaeian, M. Dehghani, and A. Diaz, "Automatic extraction of ranked SNP-phenotype associations from text using a BERT-LSTM-based method," BMC bioinformatics, vol. 24, no. 1, p. 144, 2023.